# Vehicles Swarm Intelligence: Cooperation in both Longitudinal and Lateral Dimensions *

Jia Hu, *Senior Member, IEEE*, Nuoheng Zhang, Haoran Wang, *Member*, *IEEE*, Tenglong Jiang, Junnian Zheng, and Feilong Liu

*Abstract*— Longitudinal-only platooning methods are facing great challenges on running mobility, since they may be impeded by slow-moving vehicles from time to time. To address this issue, this paper proposes a vehicles swarming method coupled both longitudinal and lateral cooperation. The proposed method bears the following contributions: i) enhancing driving mobility by swarming like a bee colony; ii) ensuring the success rate of overtaking; iii) cruising as a string of platoon to preserve sustainability. Evaluations indicate that the proposed method is capable of maneuvering a vehicle swarm to overtake slow-moving vehicles safely and successfully. The proposed method is confirmed to improve running mobility by 12.04%. Swarming safety is ensured by a safe following distance. The proposed method's influence on traffic is limited within five upstream vehicles.

## I. INTRODUCTION

Road traffic is currently facing challenges in efficiency, sustainability and safety [1]. It is reported that 81.3% of cities are suffering from traffic congestion [2]. Stop-and-go maneuvers in congested traffic increase emission by 16% and consumption by 7% [3]. Furthermore, there are 42,795 traffic fatalities in the U.S. in 2022 [4].

To handle these challenges, Cooperative Automation (CA) is proposed. A typical application of CA is the longitudinal platooning of Connected and Automated Vehicles (CAVs), represented by the Cooperative Adaptive Cruise Control (CACC) technology. By controlling CAVs to follow closely, the minimum following headway within a platoon could be reduced to 0.5 seconds [5], much smaller than the 1-2 seconds time headway of Human-driving Vehicle (HV). This reduction in following headway enhances transportation efficiency by doubling the traffic capacity to 4200 *veh/h/ln*. CACC is also confirmed to enhance driving safety and reducing fuel consumption[6], [7].

Past studies on CACC mainly utilize three types methods: feedback control method, optimal control method, and data-driven method. Linear feedforward and feedback control method is with the advantage of algorithm simplicity [8], [9]. Nevertheless, this method cannot achieve the objective of comfort and ecology. To consider multi-objectives, optimal control based method is adopted. A widely adopted method is Model Predictive Control (MPC) [10], [11]. Past studies have developed MPC-based CACC with the objective of enhancing driving safety [12], fuel efficiency [13], and driving comfort [14]. Data-driven method, such as deep reinforcement learning (DRL), has been adopted in CACC [15] to guarantee safety from system-level. Furthermore, a multi-agent reinforcement learning (MARL) framework is proposed in [16] to enhance efficiency and scalability of CACC, using decentralized training and execution simultaneously.

However, all the above CACC methods lack the lateral lane-change capability. A longitudinal-only CACC platoon may be impeded by a slow-moving vehicle from time to time. It not only reduces the driving mobility, but also cause more human takeovers. A state-of-the-art study shows that lacking lateral lane-change capability is one of the main obstacles for the commercialization of CACC [17].

Only a few state-of-the-art studies have proposed CACC controllers with lateral lane-change capability [18]. The primitive strategy is to disengage the platoon into individual CAVs for lane-change [19]. This method interrupts the continuity of platooning. CAVs may not be able to regroup into a string. To conduct CACC Lane-Change (CACCLC) maneuvers without disengagement, the Simultaneous CACCLC (Si-CACCLC) strategy is proposed [20], as depicted in Figure 1 (a). It coordinates the whole platoon to

*This paper is partially supported by National Science and Technology Major Project (No. 2022ZD0115505), National Natural Science Foundation of China (No. 52302412 and 52372317), Yangtze River Delta Science and Technology Innovation Joint Force (2023CSJGG0800), Shanghai Automotive Industry Science and Technology Development Foundation (No. 2213), Tongji Zhongte Chair Professor Foundation (No. 000000375-2018082), Shanghai Sailing Program (No. 23YF1449600), Shanghai Post-doctoral Excellence Program (No.2022571), China Postdoctoral Science Foundation (No.2022M722405), and the Science Fund of State Key Laboratory of Advanced Design and Manufacturing Technology for Vehicle (No. 32215011). *(Corresponding author: Haoran Wang)*

Jia Hu and Nuoheng Zhang are with Key Laboratory of Road and Traffic Engineering of the Ministry of Education, Tongji University, Shanghai 201804, China (e-mail: hujia@tongji.edu.cn; zhangnh@tongji.edu.cn).

Haoran Wang is with Key Laboratory of Road and Traffic Engineering of the Ministry of Education, Tongji University, Shanghai 201804, China, and State Key Laboratory of Advanced Design and Manufacturing for Vehicle Body, Hunan University, Changsha, 410082, China (e-mail: wang_haoran@tongji.edu.cn).

Tenglong Jiang is with Shanghai Motor Vehicle Inspection Certification & Tech Innovation Center Co., Ltd., No.68 South Yutian Road, Shanghai, 201805, China (e-mail: tenglongj@smvic.com.cn).

Junnian Zheng is with Hyperview Mobility (Shanghai) Co.,Ltd., No.488 Anchi Rd, Shanghai, 201805 (e-mail: junnian.zheng@hongjingdrive.com).

Feilong Liu is with Hyperview Mobility (Shanghai) Co.,Ltd., No.488 Anchi Rd, Shanghai, 201805 (e-mail: feilong.liu@hongjingdrive.com).

change lane simultaneously. Yet, this Si-CACCLC method relies on finding a sufficiently large gap to accommodate the entire platoon, making it impractical for dense traffic. To improve CACCLC capability, the Successive CACCLC (Su-CACCLC) strategy [2], [21], [22] and Making-Space CACCLC (MS-CACCLC) strategy [23], [24] is proposed, as illustrated in Figure 1 (b). They maneuver CAVs in the platoon to change lane one-by-one and cut through traffic like a snake.

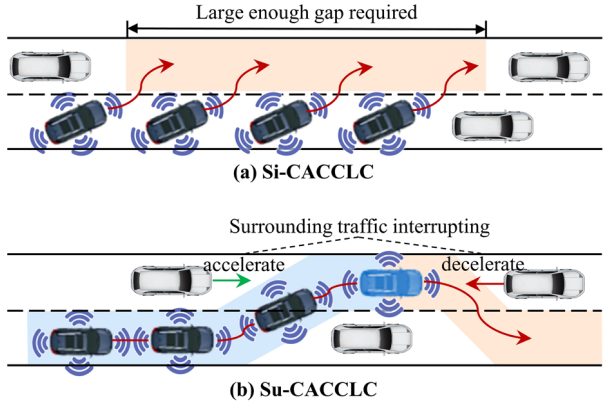

Figure 1. CCACLC methods

However, all these CACCLC methods are hardly to be implemented in actual traffic environment. Lateral movements of the whole platoon string are contingent on the yielding of surrounding HVs. Furthermore, these methods conduct a CACCLC maneuver by utilizing only one gap. This strategy is inflexible and inefficient, since utilizing more gaps on a multi-lane road would multiply accelerate the lane-change efficiency. Therefore, a more reliable and flexible lane-change method of CACC platoon is quite needed. It would be a great promotion for the field implementation of CA technology.

In this paper, we propose a vehicles swarming planner. It expands conventional longitudinal-only cooperation to longitudinal-and-lateral-coupled cooperation. The proposed planner bears the following contributions:

• **Enhancing driving mobility by swarming like a bee colony**: We propose a new cooperation scheme of multi-vehicles: vehicles swarming. It couples longitudinal and lateral cooperation together. Our proposed method enables *vehicles to swarm like a bee colony and overtake slow-moving vehicles*. Compared to the conventional longitudinal-only platooning method, driving mobility would be greatly enhanced.

• **Ensuring a successful overtaking of multi-vehicles**: Our proposed swarming method would not abandon any swarming vehicles. To conduct overtaking maneuvers successfully, our proposed method is enabled to *make space for CAVs to cut through successively*. The swarming for overtaking process would not be interrupted in real traffic.

• **Cruising as a string of platoon to preserve sustainability**: Our proposed method ensures *regrouping as a string of platoon after finishing the overtaking maneuver*.

This reserves the sustainable benefits of longitudinal platooning: reducing emissions and fuel consumption.

## II. PROBLEM FORMULATION

The objective of the proposed planner is to maneuver CAVs to overtake slow-moving HVs by swarming.

### A. System Structure

The structure of the proposed planner is presented in Figure 2. It consists of three modules:

• **Swarming maneuver planning**: This module plans sequential behaviors of all CAVs to cooperatively overtake slow-moving vehicles. It firstly grids the road segment into moving cells which are static with respect to the target slow-moving vehicle. Sequential cells are optimized for each CAV to overtake the slow-moving vehicle.

• **Trajectory generating:** This module transforms the sequential cell commands into the desired trajectory.

• **Trajectory tracking:** This module tracks the desired trajectory. It outputs the commanded acceleration and front wheel angle for each CAV.

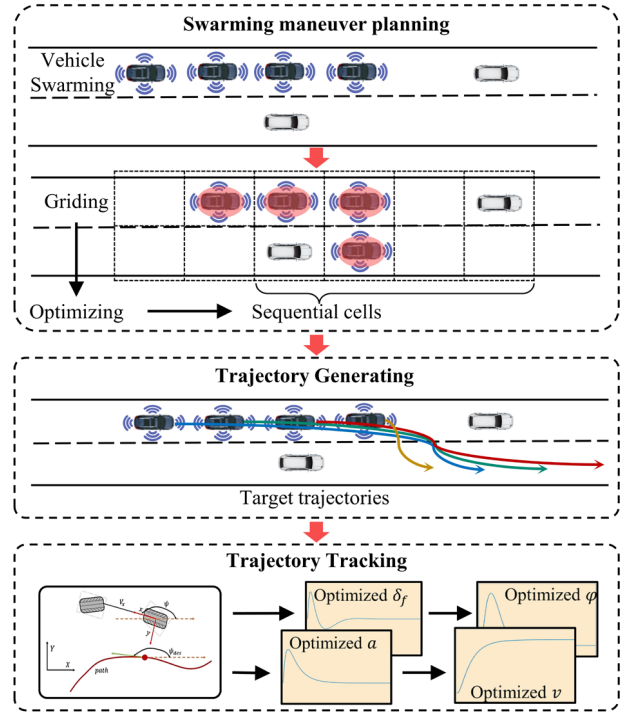

Figure 2. System structure

### B. Swarming maneuver planning

In this section, a moving-cell-based swarming maneuver planner is proposed. The sequence of desired cells is optimized for each CAV based on mixed integer quadratic programming method.

### 1) Road Segment Gridding

Gridding refers to transforming the road segment into a grid of cells. These cells are moving with a specific speed (e.g. speed of a slow-moving HV to be overtaken). A typical

gridded overtaking scenario is depicted in Figure 3. The gridding process will be elaborated as follows:

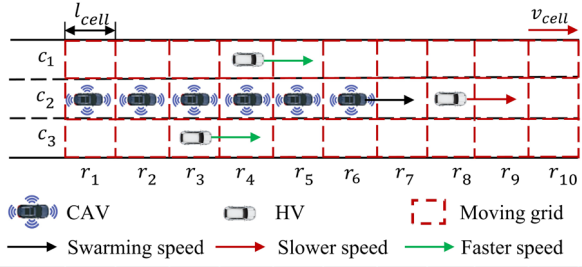

Figure 3. A typical scenario of moving grid

**Step 1.** Select the longitudinal position of the tail CAV from the swarm as the starting row of cells.

**Step 2.** Determine the size of a cell. The width of a cell is denoted as $w_{cell}$. $w_{cell}$ is set as the width of a lane. The length of a cell is denoted as $l_{cell}$. $l_{cell}$ is set as the length of a CAV $l_{CAV}$ plus the safety distance between vehicles $d_{safe}$.

**Step 3.** Determine the size of the grid. The number of columns in the grid is defined as $n_C^c$. $n_C^c$ equals to the number of lanes. The number of rows in the grid is defined as $n_C^r$. For the goal of overtaking, $n_C^r$ is set as the number of the platoon members plus the number of cells between the tail CAV and the front HV.

**Step 4.** Grid the road segment with the determined $n_C^c$ and $n_C^r$. Set the grid speed $v_{cell}$ according to the behavior of the CAV swarm. $v_{cell}$ is determined as the speed of the front HV when the swarm is executing an overtaking maneuver. It equals to a default stable speed when the swarm is in cruising mode.

*2) Moving-cell-based maneuver optimization*

With the road gridded, the planning for the CAV swarm is transformed into the task of shifting CAVs within the grid. On the basis, it is modeled as a mixed integer programming problem in this section.

*(1) Variable definition*

The temporal domain is discretized into time steps $k \in [1, N]$ with a step length of $\Delta t$, where $N$ is the total steps of planning horizon. The number of CAVs is denoted as $n_{CAV}$, and the number of HVs within the grid is denoted as $n_{HV}$.

At the $k$-th step, the occupancy states, indicating whether the $i$-th CAV is in the $p$-th row and whether the $i$-th CAV is in the $q$-th column, are defined as follows:

$$r_p^{i,k} = \begin{cases} 1 & the\ i-th\ CAV\ is\ in\ the\ p-th\ row \\ 0 & the\ i-th\ CAV\ is\ not\ in\ the\ p-th\ row \end{cases} \quad (1)$$

$$c_q^{i,k} = \begin{cases} 1 & the\ i-th\ CAV\ is\ in\ the\ q-th\ column \\ 0 & the\ i-th\ CAV\ is\ not\ in\ the\ q-th\ column \end{cases} \quad (2)$$

where $i \in [1, n_{CAV}]$, $p \in [1, n_C^r]$, $q \in [1, n_C^c]$. Equation (1)-(2) define the variables to be optimized, which constitute the target sequential cells for each CAV. Similarly, the occupancy states of the $j$-th HV are defined as follows:

$$r_p^{j,k} = \begin{cases} 1 & the\ j-th\ HV\ is\ in\ the\ p-th\ row \\ 0 & the\ j-th\ HV\ is\ not\ in\ the\ p-th\ row \end{cases} \quad (3)$$

$$c_q^{j,k} = \begin{cases} 1 & the\ j-th\ HV\ is\ in\ the\ q-th\ column \\ 0 & the\ j-th\ HV\ is\ not\ in\ the\ q-th\ column \end{cases} \quad (4)$$

where $i \in [1, n_{HV}]$, $p \in [1, n_C^r]$, $q \in [1, n_C^c]$. During the planning horizon, states of HVs within the grid are assumed to be predictable, thus $r_p^{j,k}$ and $c_q^{j,k}$ are known constants.

*(2) Objective function*

For the purpose of overtaking the slow-moving HV, the cost function is formulated as follows:

$$\begin{aligned} \min_{r,c} &\sum_{k=1}^{N}\sum_{i=1}^{n_G}\sum_{p=1}^{n_C^r-n_G^r} W_{tar}\left(r_p^{i,k}\right)^2 + \\ &\sum_{k=1}^{N-1}\sum_{i=1}^{n_G}\sum_{p=1}^{n_C^r} W_{lon}\left(r_p^{i,k+1} - r_p^{i,k}\right)^2 + \\ &\sum_{k=1}^{N-1}\sum_{i=1}^{n_G}\sum_{q=1}^{n_C^c} \delta W_{lat}\left(c_q^{i,k+1} - c_q^{i,k}\right)^2 + \\ &\sum_{k=1}^{N-1}\sum_{i=1}^{n_G}\sum_{q=1}^{n_C^c} (1-\delta) W_{lat}\left(l_{index} - c_q^{i,k}\right)^2 \end{aligned} \quad (5)$$

where $W_{tar}$ is the weight coefficient of the travelling forward target, $W_{lon}$ is the weight coefficient of the longitudinal moving cost, $W_{lat}$ is the weight coefficient of the lateral moving cost.

Three cost components are included in Equation (5). The first part is the cost of travelling forward target, which drives the swarm to progress in the grid. Only if a CAV of the swarm arrives at the foremost cell, i.e., the row numbered $n_G^r$, can this part be minimized. The second part and the third part represent the longitudinal and lateral moving costs, respectively. Both are designed to prevent erroneous planning, such as two CAVs switch their position or one CAV oscillates laterally within a single step. The fourth part aims to guide CAVs to regroup into a platoon in the lane $l_{index}$ after overtaking the front HV. A 0-1 integer decision variable $\delta$ is defined in this part. It equals to 1 when there are still CAVs impeded by the front HV, and 0 otherwise.

*(3) Constraints*

The movement of all vehicles in the grid must adhere to the laws of motion and the conditions of consistency. Therefore, CAVs are limited by the following constraints.

• **A CAV only occupies one cell**: According to physical conditions such as the vehicle length, a CAV of the swarm can only occupy a cell at each step. Therefore, for $\forall i \in [1, n_G]$ and $\forall k \in [1, N]$:

$$\sum_{p=1}^{n_C^r} r_p^{i,k} = 1 \quad (6)$$

$$\sum_{q=1}^{n_C^c} c_q^{i,k} = 1 \quad (7)$$

• **Rules for CAV's moving**: At each step, CAVs can only move between adjacent cells. It avoids moving across the grid longitudinally, laterally or diagonally.

*Longitudinally move less than one cell within one step*: For $\forall i \in [1, n_{CAV}]$ and $\forall k \in [1, N-1]$:

$$r_{p_1}^{i,k} + r_{p_2}^{i,k+1} \leq 1 \quad (8)$$

where $p_1 \in [1, n_C^r]$, $p_2 \in [1, n_C^r]$. $p_1$ and $p_2$ are two non-adjacent ordered row numbers.

*Laterally move less than one cell within one step*: For $\forall i \in [1, n_{CAV}]$ and $\forall k \in [1, N-1]$:
$$c_{q_1}^{i,k} + c_{q_2}^{i,k+1} \leq 1 \quad (9)$$
where $p \in [1, n_C^r]$, $q \in [1, n_C^c]$. $p_1$ and $p_2$ are two non-adjacent ordered column numbers.

*Cornerwise moving not allowed*: For $\forall i \in [1, n_{CAV}]$ and $\forall k \in [1, N-1]$:
$$r_p^{i,k} + r_{p+1}^{i,k+1} + c_q^{i,k} + c_{q+1}^{i,k+1} \leq 3 \quad (10)$$
$$r_{p+1}^{i,k} + r_p^{i,k+1} + c_{q+1}^{i,k} + c_q^{i,k+1} \leq 3 \quad (11)$$
$$r_{p+1}^{i,k} + r_p^{i,k+1} + c_q^{i,k} + c_{q+1}^{i,k+1} \leq 3 \quad (12)$$
$$r_p^{i,k} + r_{p+1}^{i,k+1} + c_{q+1}^{i,k} + c_q^{i,k+1} \leq 3 \quad (13)$$
where $p \in [1, n_C^r - 1]$, $q \in [1, n_C^c - 1]$.

- **Collision avoiding constraints**: In order to avoid collisions, CAVs in the grid are not allowed to occupy a cell at the same time in all planning steps. Therefore, for $\forall k \in [2, N]$, $\forall p \in [1, n_C^r]$, $\forall q \in [1, n_C^c]$, $\forall i_1 \in [1, n_G]$, $\forall i_2 \in [1, n_G]$:
$$r_p^{i_1,k} + r_p^{i_2,k} + c_q^{i_1,k} + c_q^{i_2,k} \leq 3 \quad (14)$$
where $i_1$ and $i_2$ are the different numbers of CAVs, $p$ is the row number of the grid with $p \in [1, n_C^r - 1]$, $q$ is the column number of the grid with $q \in [1, n_C^c - 1]$.

Meanwhile, CAVs and HVs in the grid are not allowed to be in a same cell within a single step. Hence, for $\forall k \in [2, N]$, $\forall i \in [1, n_G]$, $\forall j \in [1, n_{HV}]$:
$$r_{p_{j,k}}^{i,k} + c_{q_{j,k}}^{i,k} \leq 1 \quad (15)$$
where $i$ is the number of a CAV, $p_{j,k}$ is the row number of the $j$-th HV at step $k$, and $q_{j,k}$ is the column number of the $j$-th HV at step $k$.

- **Space making constraints**: The gaps for lane-change may be narrowed by other HVs during the overtaking process. To address this, the proposed vehicles swarming planner assigns several CAVs in the front to change lanes, slow HVs down and make enough space. The number of assigned CAVs and the lane-change target are determined based on the lanes where HVs are detected. The assignment is formulated as space making constraints in Equation (16) and (17):
$$c_{q_{j,k}}^{i,k+1} + \varepsilon_k \geq 1 \quad (16)$$
$$\varepsilon_k (\sum_{p=p_{i,k}}^{p_{i,k}} r_p^{i,k+1} - 1) = 0 \quad (17)$$
where $k$ is the step when HVs are detected in other lanes, $\varepsilon_k$ is a binary variable which equals to 1 when HVs are detected in other lanes and equals to 0 otherwise, $q_{j,k}$ is the column number of the detected $j$-th HV at step $k$, $p_{j,k}$ is the row number of the detected $j$-th HV at step $k$, and $p_{i,k}$ is the row number of the assigned $i$-th CAV at step $k$.

- **Initial condition definition**: At the first step, initial positions of all CAVs are available. Hence for $k = 1$, $\forall i \in [1, n_{CAV}]$, $\forall p \in [1, n_C^r]$ and $\forall q \in [1, n_C^c]$:
$$r_p^{i,1} = \begin{cases} 1 & \text{the } i-\text{th CAV is in the } p-\text{th row} \\ 0 & \text{the } i-\text{th CAV is not in the } p-\text{th row} \end{cases} \quad (18)$$

$$c_q^{i,1} = \begin{cases} 1 & \text{the } i-\text{th CAV is in the } q-\text{th column} \\ 0 & \text{the } i-\text{th CAV is not in the } q-\text{th column} \end{cases} \quad (19)$$

*(4) Planning update scheme*

The proposed swarming maneuver planner is updated triggered by specific events. As illustrated in Figure 4, the CAV swarm executes the planned control vector after the initial loop of planning. The planner will be updated when either of the two conditions is met: new HVs enter the grid, or the planned control vectors are fully executed.

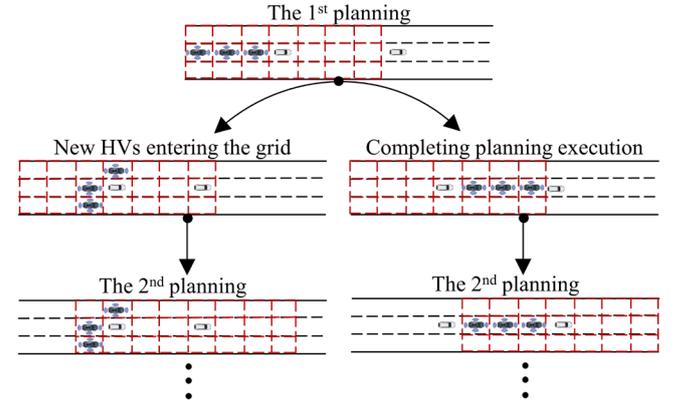

Figure 4. Event-triggered planning update scheme

## C. Trajectory Generating

In this section, the target sequential cells commands acquired from the swarming maneuver planner are transformed into the desired trajectory for each CAV. Given the step length $\Delta t$, the length of the cell $l_{cell}$ and the speed of the grid $v_{cell}$, the trajectories are generated according to Equation (20)-(22).

$$t_k = t_0 + k\Delta t \quad (20)$$
$$s_k^i = s_0 + v_{cell} k \Delta t + \frac{1}{2} l_{cell} + (row_k^i - 1) l_{cell} \quad (21)$$
$$y_k^i = \begin{cases} 3 & col_k^i = 3 \\ 0 & col_k^i = 2 \\ -3 & col_k^i = 1 \end{cases} \quad (22)$$

where $k$ is the step number, $t_k$ is the absolute time, $t_0$ is the initial time when the overtaking starts, $s_k^i$ is the absolute longitudinal position of the $i$-th CAV at step $k$, $s_0$ is the starting longitudinal position of the grid when the overtaking process begins, $row_k^i$ is the row number of the $i$-th CAV position at step $k$, $y_k^i$ is the absolute lateral position of the $i$-th CAV at step $k$, and $col_k^i$ is the column number of the $i$-th CAV position at step $k$.

## D. Trajectory Tracking

In this section, an optimal-control-based trajectory tracking method is proposed. It is designed to track the trajectories generated by the upper module, and subsequently produce target acceleration and target front wheel angle for each CAV in the swarm.

The trajectory tracking method is formulated in the mixed domain. The longitudinal trajectories are tracked based on the temporal domain. With the input of the tracked longitudinal trajectories, the lateral trajectories are then tracked in the spatial domain. The goal of tracking for a CAV is to reaching the corresponding lateral position when it arrives at a given longitudinal position. This approach decouples longitudinal and lateral tracking errors, thereby enhancing tracking accuracy.

*1) State Definition*

The trajectory tracking method is formulated in the reference path coordinate system. As shown in Figure 5, the longitudinal direction is defined along the reference path, The lateral direction is defined to be perpendicular to the reference path.

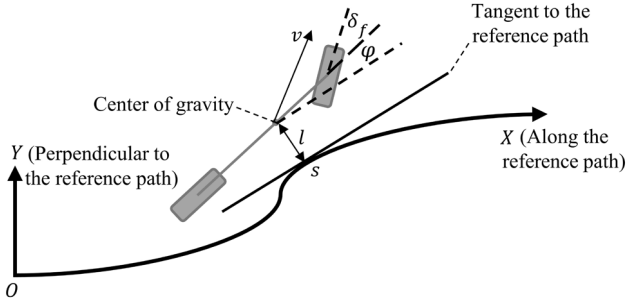

Figure 5. The reference path coordinate system

Based on the mixed domain, the longitudinal system state vector $X_k^{lon}$, the lateral system state vector $X_k^{lat}$, the longitudinal control vector $U_k^{lon}$ and the lateral control vector $U_k^{lat}$ are defined as follows:

$$X_k^{lon} \stackrel{\text{def}}{=} [s_k, v_k] \quad (23)$$
$$X_k^{lat} \stackrel{\text{def}}{=} [l_k, \varphi_k] \quad (24)$$
$$U_k^{lon} \stackrel{\text{def}}{=} a_k \quad (25)$$
$$U_k^{lat} \stackrel{\text{def}}{=} \delta_k \quad (26)$$

where $s_k$ is the longitudinal position at step $k$, $v_k$ is the speed at step $k$, $l_k$ is the tracking error of lateral position at step $k$, $\varphi_k$ is the yaw angle relative to the tangent of the reference path at step $k$, $a_k$ is the acceleration at step $k$, and $\delta_k$ is the front wheel angle at step $k$.

In order to minimize the tracking error, the desired longitudinal state $X_{des}^{lon}$ and the desired lateral state $X_{des}^{lat}$ are defined as follows:

$$X_{des}^{lon} \stackrel{\text{def}}{=} [s_{des}, v_{des}] \quad (27)$$
$$X_{des}^{lat} \stackrel{\text{def}}{=} [0,0] \quad (28)$$

where $s_{des}$ is the desired longitudinal position and $v_{des}$ is the desired speed.

*2) System Dynamics*

The longitudinal system dynamics and lateral system dynamics are modeled into Equation (29)-(35):

$$X_{k+1}^{lon} = A^{lon} X_k^{lon} + B^{lon} U_k^{lon}, k \in [0, K-1] \quad (29)$$

$$X_{k+1}^{lat} = A^{lat} X_k^{lat} + B^{lat} U_k^{lat} + C_k^{lat}, k \in [0, K-1] \quad (30)$$

with

$$A^{lon} = \Delta t \times \begin{bmatrix} 0 & 1 \\ 0 & 0 \end{bmatrix} + I_{2 \times 2} \quad (31)$$

$$B^{lon} = \Delta t \times \begin{bmatrix} 0 \\ 1 \end{bmatrix} \quad (32)$$

$$A^{lat} = \Delta s \times \begin{bmatrix} 0 & 1 \\ 0 & 0 \end{bmatrix} + I_{2 \times 2} \quad (33)$$

$$B^{lat} = \Delta s \times \begin{bmatrix} 0 \\ \frac{1}{\omega} \end{bmatrix} \quad (34)$$

$$C_k^{lat} = \Delta s \times \begin{bmatrix} 0 \\ -\kappa_k \end{bmatrix} \quad (35)$$

where $\Delta t$ is the time increment, $\Delta s$ is the distance increment, $\omega$ is the wheelbase of the CAV, and $\kappa_k$ is the curvature of the reference path at step $k$.

*3) Trajectories Tracking Model*

The control objective is to minimize the state errors while balancing the control costs. Hence the cost function can be formulated as the sum of squared state errors and the squared value of the control vector.

The longitudinal planning model is designed as follows:

$$J^{lon} = \underbrace{\frac{1}{2}(X_k^{lon} - X_{des}^{lon})^T Q^{lon}(X_k^{lon} - X_{des}^{lon})}_{\text{state error cost}} + \underbrace{\frac{1}{2} r^{lon}(U_k^{lon})^2)}_{\text{control cost}} \quad (36)$$

$$s.t. X_{k+1}^{lon} = A^{lon} X_k^{lon} + B^{lon} U_k^{lon}, k \in [0, K-1] \quad (37)$$

$$v_{min} \leq v \leq v_{max} \quad (38)$$

$$a_{min} \leq a \leq a_{max} \quad (39)$$

with

$$Q^{lon} = \begin{bmatrix} q_s & 0 \\ 0 & q_v \end{bmatrix} \quad (40)$$

where $J^{lon}$ is the longitudinal cost function, $Q^{lon}$ is the longitudinal weight matrix, $q_s$ is the coefficient of the longitudinal position weight, $q_v$ is the coefficient of the longitudinal speed weight, and $r^{lon}$ is the coefficient of the acceleration cost weight.

The lateral planning model is designed as follows:

$$J^{lat} = \underbrace{\frac{1}{2}(X_k^{lat} - X_{des}^{lat})^T Q^{lat}(X_k^{lat} - X_{des}^{lat})}_{\text{state error cost}} + \underbrace{\frac{1}{2} r^{lat}(U_k^{lat})^2)}_{\text{control cost}} \quad (41)$$

$$s.t. X_{k+1}^{lat} = A^{lat} X_k^{lat} + B^{lat} U_k^{lat} + C_k^{lat}, k \in [0, K-1] \quad (42)$$

$$\delta_{min} \leq \delta \leq \delta_{max} \quad (43)$$

with

$$Q^{lat} = \begin{bmatrix} q_l & 0 \\ 0 & q_\varphi \end{bmatrix} \quad (44)$$

where $J^{lat}$ is the lateral cost function, $Q^{lat}$ is the lateral weight matrix, $q_l$ is the weight of the lateral position error, $q_\varphi$ is the weight of the yaw angle, and $r^{lat}$ is the weight of the front wheel angle.

III. SOLUTION METHOD

*A. Solution to the Vehicles Swarming Maneuver Planner*

The proposed swarming maneuver planner is formulated as a mixed integer programming problem. A high-performance mathematical programming solver, Gurobi

10.0.1, is used to solve the problem. The outputs are occupancy state variables, which constitute the target sequential cells of CAVs within the grid.

*B. Solution to the trajectory tracking method*

The proposed trajectory tracking method is solved via the dynamic programming algorithm previously developed by this research group [2], [23], [25]. It transforms the optimal control problem into multiple overlapping sub-problems, then solves each sub-problem step by step. The solving process is detailed as follows:

**Input:** initial states $X_0$, $X_{des}$, $Q$, $R$.

***Step 1.*** For the terminal step $k = K$:
$$\widetilde{Q}_k = Q \tag{45}$$
$$\widetilde{D}_k = 0 \tag{46}$$
$$\widetilde{E}_k = 0 \tag{47}$$

***Step 2.*** Calculate concomitant matrices backwardly.
For $k = K-1, \cdots, 0$:
$$\widetilde{Q}_k = G_k^T R G_k + S_k^T \widetilde{Q}_{k+1} S_k + Q \tag{48}$$
$$\widetilde{D}_k = G_k^T R H_k + S_k^T \widetilde{Q}_{k+1} T_k + S_k^T \widetilde{D}_{k+1} \tag{49}$$
$$\widetilde{E}_k = \frac{1}{2} H_k^T R H_k + \frac{1}{2} T_k^T \widetilde{Q}_{k+1} T_k \tag{50}$$
with
$$P_k = (R + B^T \widetilde{Q}_{k+1} B)^{-1} \tag{51}$$
$$G_k = -P_k B^T \widetilde{Q}_{k+1} A \tag{52}$$
$$H_k = -P_k B^T \widetilde{D}_{k+1} \tag{53}$$
$$S_k = A + BG_k \tag{54}$$
$$T_k = BH_k \tag{55}$$

***Step 3.*** Calculate control vector and state vector forwardly.
For $k = 0, 1, \cdots, K-1$:
$$U_k = G_k X_k + H_k \tag{56}$$
$$X_{k+1} = S_k X_k + T_k \tag{57}$$

**Output:** optimal control vector $U_k$ and state vector $X_k$ of each step.

## IV. EVALUATION

The proposed vehicles swarming method has been evaluated in the following aspects: i) function validation; ii) mobility; iii) safety; iv) influence on traffic.

*A. Experiment Design*

A simulation is conducted on a numerical simulation platform based on MATLAB. In the experiment, longitudinal position, lateral position, speed, and acceleration of CAVs and HVs are collected for analysis.

Under control of the proposed method, the platoon cruises in the context of traffic on a three-lanes highway. In the context of traffic, the platoon may be impeded by a slow-moving HV, as shown in Figure 6. To ensure cruising mobility, overtaking is a must. Furthermore, to ensure a successful overtaking maneuver, the swarming controller shall be capable of blocking the left-rear and right-rear HV to make space for CAVs to cut through.

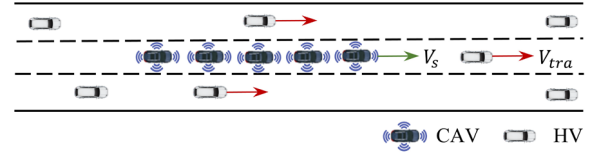

Figure 6. Scenario of interest

*B. Controller Types*

There are three types of controllers in the simulation:

- *Proposed vehicles swarming controller*: The proposed vehicles swarming controller maneuvers CAVs to swarm and overtake slow-moving HVs.
- *Baseline longitudinal-only CACC controller*: The baseline CACC controller can only control CAVs to follow slowing-moving vehicles without overtaking capability.
- *Human driving controller:* All surrounding HVs are controlled by a widely used Intelligent Driver Model (IDM) [26].

*C. Parameter Setting*

Parameters of the experiment are set as follows:
- Lane width $w_{lane}$: 3m;
- Vehicle length $l_v$: 5m;
- Number of CAVs $n_{cav}$: 6;
- Time interval $\Delta T$: 3s;
- Cell length $l_{cell}$: 5m;
- Time step $\Delta t$: 0.03s;
- Swarm cruising speed $v_s$: 20m/s;
- Speed of HVs $v_{tra}$: 17.5m/s;
- Minimum safety distance between CAVs: 5m;
- Desired safety distance between CAVs: 10m;
- Minimum car following headway: 1.2s;
- Initial following distances of the upstream HVs: 25m;
- Range of acceleration: [-4$m/s^2$, 3$m/s^2$];
- Range of front wheel angle: [-450°, 450°];
- The parameters of IDM are set according to [27].

*D. Measure of Effectiveness*

Measures of Effectiveness (MOEs) are designed as follows:

- **Function validation:** The function of CAVs swarming for overtaking is verified by vehicle trajectories.
- **Mobility:** This MOE is qualified by the speed (average value, extreme value and variance) and travel time of the platoon.
- **Safety:** This MOE is qualified by the following distance of CAVs.
- **Influence on traffic:** The efficiency and safety of upstream traffic is evaluated by speed, following distances and vehicle trajectories.

*E. Experiment Results*

Results demonstrate that the proposed vehicles swarming method: i) is capable of maneuvering CAVs to overtake slow-moving vehicles safely and successfully; ii) improves running mobility by 12.04% compared to the baseline CACC controller; iii) maintains a safe following distance; iv) limits its influence on traffic within four following HVs.

*1) Function validation*

The proposed vehicles swarming controller is confirmed with the capability of maneuvering CAVs in platoon to overtake slow-moving vehicles by swarming like a bee colony. In the scenario of interest, the sequence of desired cells is shown in Figure 7. The platoon is planned to overtake the slow-moving HV within 14 behaving steps. The overtaking process utilizes both the left and right lanes. Two CAVs are selected to block the HVs on the left and right lane, in order to ensure a stable cut-through gap.

Utilizing the planned strategy in Figure 7, CAVs' execution trajectories are shown in Figure 8. All designed functions are verified, including blocking HVs, making space, swarming to overtake, cutting through successively and regrouping into a platoon. At time $0s$, the platoon makes an overtaking decision. The leading CAV (green) and the second CAV (yellow) are assigned to block the HVs on the left and right lane. At time $3s$, the green and yellow CAVs are making spaces for the following CAVs to cut through. From time $9s$ to $27s$, all following CAVs cut through the gap successively and swarm to overtake the slow-moving HV. From time $27s$ to $42s$, CAVs reform into a platoon and continue the cruising.

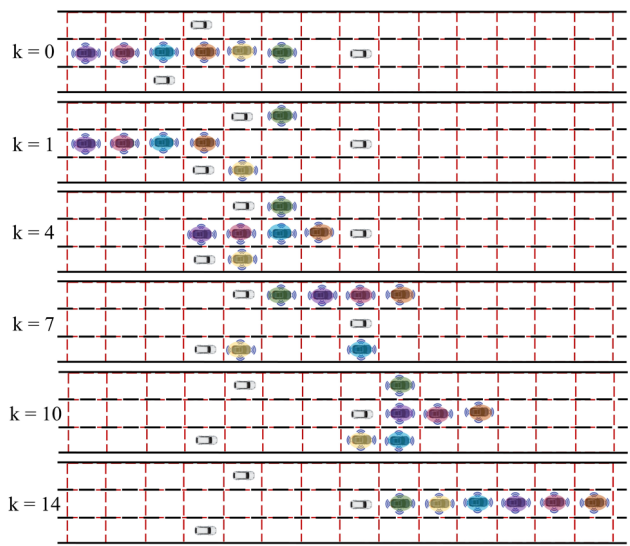

Figure 7. Vehicles swarming planning results

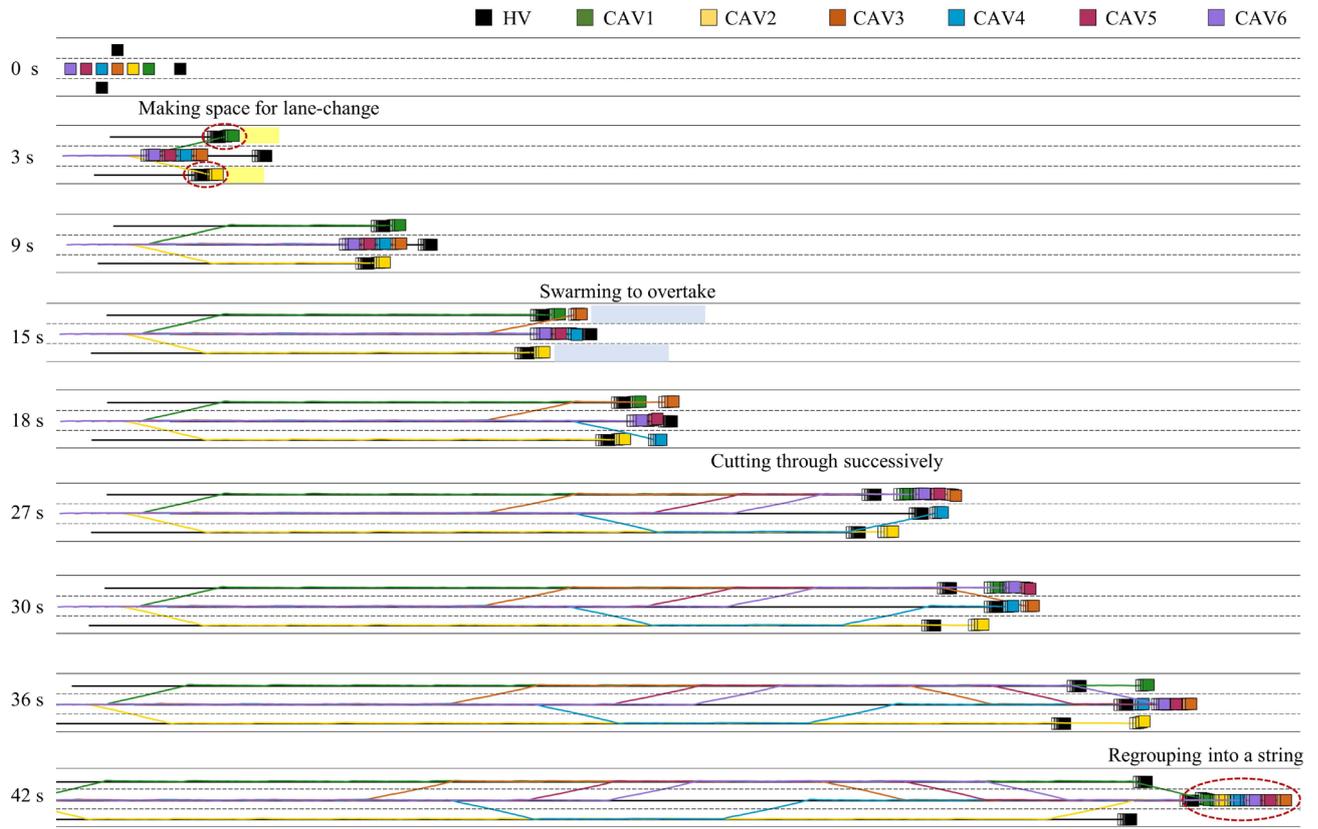

Figure 8. Vehicle trajectories of swarming

*2) Mobility*

The proposed vehicles swarming controller is confirmed to enhance platoon's mobility compared to the baseline CACC controller. As shown in Figure 9, the proposed controller ensures the whole platoon to travel at an average speed of 19.43*m/s*, which is 12.04% greater than the baseline CCAC controller. This confirms that the overtaking capability reduces traveling delay caused by slow-moving HVs.

*3) Safety*

The safety of vehicles swarming maneuver is confirmed by CAVs' following distance, as illustrated in Figure 10. Results show that all CAVs maintain a safe following distance. The average following distance of CAVs is 16.80*m*, which is 68% greater than the desired safe distance. The minimum following distance is also greater than the safe distance (set as 5 meters for CAVs)

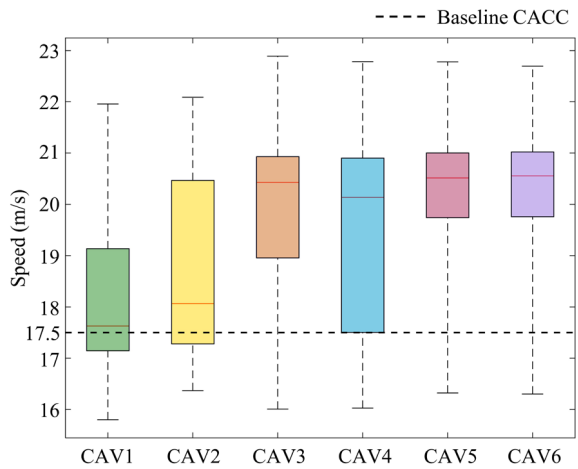

Figure 9. Average speed of CAVs in the platoon

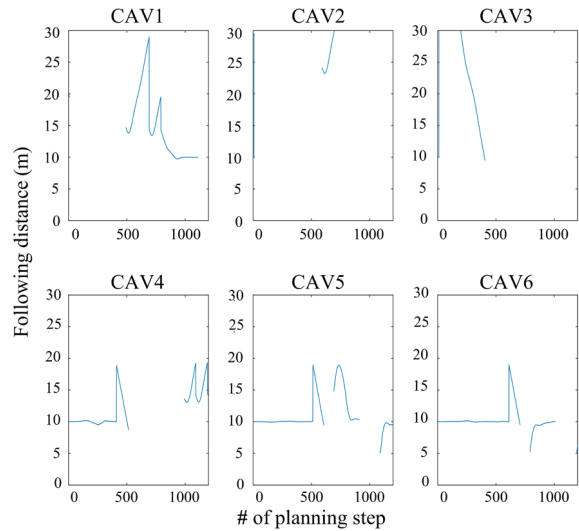

Figure 10. Following distance of CAVs

*4) Influence on traffic*

The influence of vehicles swarming on upstream traffic is evaluated by the simulation in the context of traffic. Vehicles trajectories are illustrated in Figure 11. In the CAVs swarming and overtaking process, the behavior of blocking HVs may decelerate the upstream traffic. However, since the blocking is gently and smoothly, the stability of upstream traffic would not be deteriorated to a large extent.

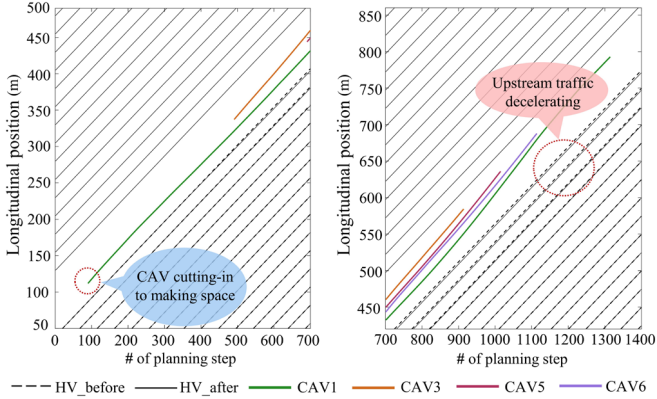

Figure 11. Traffic trajectories on the left lane when CAVs swarm

The speed trajectories of upstream HVs are illustrated in Figure 12 (a). The proposed vehicles swarming method indeed causes a few upstream vehicles to slow down. The first following HV's speed is reduced by 2.19%. The second following HV's speed is reduced by 1.51%. The third following HV's speed is reduced by 0.92%. The forth following HV's speed is reduced by 0.42%. It could be found that the speed reduction rate decreases along the traffic stream. The speed oscillation phenomenon fades away after the fourth following HV. This result demonstrates that the efficiency influence of the proposed method on upstream traffic is limited and insignificant.

The following distance of upstream traffic is illustrated in Figure 12 (b). It shows that the minimal following distance is 23.87$m$, equaling to a time headway of 1.4$s$ at the speed of 17.15$m/s$. Since the minimum car following headway is set as 1.2$s$, upstream HVs are in a safe following status. When the overtaking process finishes, the following distance restores stability.

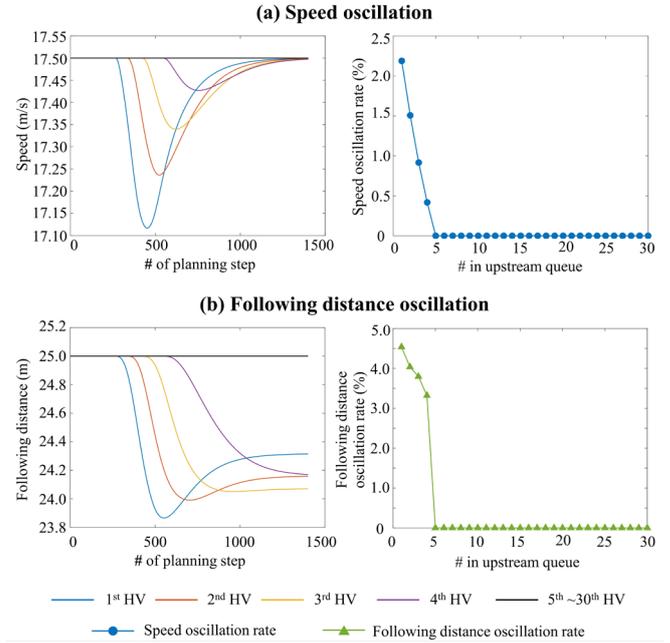

Figure 12. Safety and efficiency impact on the upstream traffic

## V. CONCLUSION AND FUTURE RESEARCH

This paper proposes a vehicles swarming method coupled both longitudinal and lateral cooperation. The proposed method bears the following contributions: i) enhancing driving mobility by swarming like a bee colony; ii) ensuring a successful overtaking of multi-vehicles; iii) cruising as a string of platoon to preserve sustainability. Evaluation results indicate that:

• The proposed method improves the mobility by 12.04% compared to the longitudinal-only CCAC controller.
• The proposed method makes sure the safety of swarming by maintaining a following distance 68% greater than the minimum distance.
• The proposed method only influences four upstream vehicles with 1.26% of speed oscillations on average. Furthermore, upstream traffic would return to a stable status after the overtaking maneuver.

In the present research, the proposed method assumes that the states of background vehicles do not undergo rapid changes within a short prediction period. Future research could concentrate on the development of more precise methods for predicting the states of background vehicles.

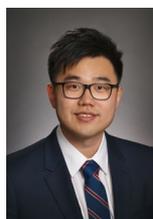

**Jia Hu** (Member, IEEE) is currently working as a Zhongte Distinguished Chair of Cooperative Automation with the College of Transportation Engineering, Tongji University. Before joining Tongji University, he was a Research Associate with the Federal Highway Administration (FHWA), USA. He is an Editorial Board Member of the *Journal of Intelligent Transportation Systems* and the *International Journal of Transportation Science and Technology*. He is a member of TRB (a Division of the National Academies) Vehicle Highway Automation Commit-tee, the Freeway Operations Committee, Simulation subcommittee of Traffic Signal Systems Committee, and the Advanced Technologies Committee of the ASCE Transportation and Development Institute. He is the Chair of the Vehicle Automation and Connectivity Committee of the World Transport Convention. He is an Associate Editor of the American Society of Civil Engineers *Journal of Transportation Engineering* and IEEE OPEN JOURNAL OF INTELLIGENT TRANSPORTATION SYSTEMS.

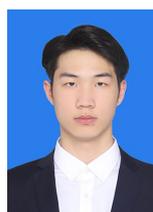

**Nuoheng Zhang** received the bachelor's degree in traffic engineering from Jilin University, Changchun, China, in 2023. He is currently pursuing the master's degree with Tongji University. His main research interests include cooperative automation, optimal control, decision making and behavioral planning.

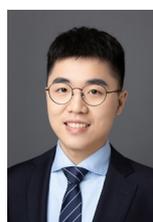

**Haoran Wang** received the bachelor's degree in transportation engineering from Tongji University, Shanghai, China, in 2017, and the Ph.D. degree from Tongji University in 2022. He is currently a Postdoctoral Researcher with the College of Transportation Engineering, Tongji University. He is a researcher on vehicle engineering, majoring in intelligent vehicle control and cooperative automation. Dr. Wang served the IEEE TRANSACTIONS ON INTELLIGENT VEHICLES, IEEE TRANSACTIONS ON INTELLIGENT TRANSPORTATION SYSTEMS, *Journal of Intelligent Transportation Systems*, and IET *Intelligent Transport Systems* as peer reviewers with a good reputation.

**Tenglong Jiang** was born in Qingdao, China. He received the M.S. degree in vehicle engineering from Tongji University, Shanghai, China, in 2022. He

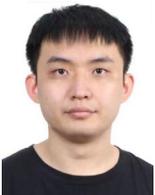
is currently employed at the Shanghai Motor Vehicle Inspection Certification & Tech Innovation Center Co., Ltd. His research interests include connected and automated vehicle, simulation test, ITS, and vehicle dynamics.

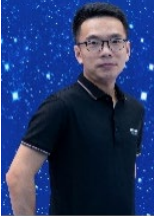
**Junnian Zheng** was born in Shanghai, China. He holds B.S. degree in mechanical engineering from Shanghai Jiao Tong University, and M.S. and Ph.D. degrees in mechanical engineering from Texas A&M University. He is currently the director of innovation and advanced engineering at Hyperview Mobility (Shanghai). His research interests include ADAS and autonomous driving system for passenger and commercial vehicles, embodied AI, and large language models for self-driving.

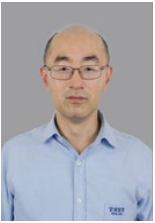
**Feilong Liu** was born in Beijing, China. He received his PhD in Electrical Engineering from the University of Cambridge in 2006. He is currently employed at the Hyperview Mobility (Shanghai) Co.,Ltd. His research interests include ADAS and autonomous driving system for passenger and commercial vehicles, embodied AI, ITS, and vehicle dynamics.